\documentclass[conference]{IEEEtran}
\IEEEoverridecommandlockouts
\pdfoutput=1
\usepackage{cite}
\usepackage{amsmath,amssymb,amsfonts,amsthm}
\usepackage{algorithmic}
\usepackage{graphicx}
\usepackage[all]{xy}
\usepackage{textcomp}
\usepackage{xcolor}
\def\BibTeX{{\rm B\kern-.05em{\sc i\kern-.025em b}\kern-.08em
    T\kern-.1667em\lower.7ex\hbox{E}\kern-.125emX}}
\begin{document}

\title{Large scale classification in deep neural network with Label Mapping\\
}

\author{\IEEEauthorblockN{1\textsuperscript{st} Qizhi Zhang}
\IEEEauthorblockA{\textit{Alibaba Group} \\
Hangzhou, China \\
qizhi.zqz@alibaba-inc.com}
\and
\IEEEauthorblockN{2\textsuperscript{nd} Kuang-Chih Lee}
\IEEEauthorblockA{\textit{Alibaba Group} \\
Hangzhou, China \\
kuang-chih.lee@alibaba-inc.com}
\and
\IEEEauthorblockN{3\textsuperscript{rd} Hongying Bao}
\IEEEauthorblockA{\textit{Alibaba Group} \\
Hangzhou, China \\
hongying.bhy@alibaba-inc.com} \\
\and
\IEEEauthorblockN{4\textsuperscript{th} Yuan You}
\IEEEauthorblockA{\textit{Alibaba Grooup} \\
Hangzhou, China \\
youyuan.yy@alibaba-inc.com}
\and
\IEEEauthorblockN{5\textsuperscript{th} Wenjie Li}
\IEEEauthorblockA{\textit{Alibaba Group} \\
Hangzhou, China \\
wenjie.lwj@alibaba-inc.com}
\and
\IEEEauthorblockN{5\textsuperscript{th} Dongbai Guo}
\IEEEauthorblockA{\textit{Alibaba Group} \\
Hangzhou, China \\
dongbai.gdb@alibaba-inc.com}
}

\maketitle

\begin{abstract}
   In recent years, deep neural network is widely used in machine learning.
The  multi-class classification problem
is a class of important problem in machine learning. 
However, in order to solve those 
types of multi-class classification problems effectively, 
the required network size should have hyper-linear growth with respect to the number of classes. 
Therefore, it is infeasible to solve the multi-class classification 
problem using deep neural network when the number of classes are huge.
This paper presents a method, so called Label Mapping (LM), to solve this problem by
decomposing the original classification problem to several smaller 
sub-problems which are solvable theoretically. Our method is an ensemble method like
error-correcting output codes (ECOC), but it allows base learners to be multi-class classifiers with  
different number of class labels. 
We propose two design principles for LM, one is to maximize the 
number of base classifier which can separate two different classes, and the other 
is to keep all base learners to be independent as possible in order to reduce the redundant information. 
Based on these principles, two different LM algorithms are derived using number theory and information theory. Since each base learner can be trained independently, it is easy to scale our method 
into a large scale training system. Experiments show that our proposed method outperforms the standard 
one-hot encoding and ECOC significantly in terms of accuracy and model complexity.
\end{abstract}


 \newtheorem{lem}{Lemma}
 \newtheorem{cor}[lem]{Corollary}
 \newtheorem{thm}[lem]{Theorem}
 \newtheorem{prop}[lem]{Proposition}
 \theoremstyle{definition}
 \newtheorem{defn}[lem]{Definition}
 \theoremstyle{remark}
 \newtheorem{rem}[lem]{Remark}
 \newtheorem{exa}[lem]{Example}

\newcommand{\norm}[1]{\parallel\! #1 \!\parallel}
\newcommand{\mnorm}[1]{$\parallel\! #1 \!\parallel$}
\section{Introduction}

Deep learning has become one of the major research areas in the machine learning community. One of the challenge is that 
the structure of the deep network model is usually complicated. 
For general multi-class classification problems, the required parameters of the deep network need to have hyper-linear growth with 
respect to the class number. If the number of classes are large, the classification problem will become infeasible 
because the required resources for model computation and storage will be huge. However, 
today there are lots of applications that require to perform classification with huge number of classes, 
such as language model of word level, image recognition of shopping items in e-commerce (multi-billions of shopping items today in Taobao and Amazon), 
as well as handwriting recognition of 10K Chinese characters.

In fact, A general deep neural network classifier of $N$ classes can be treated as a series connection of a complex embedding
 in Euclidean space to the
last but one layer, and a softmax classifier softmax$(Ax+b)$ of $N$ classes in the last layer. 
The complex embedding can be 
interpreted as a clustering process to cluster data based on their class labels, and the last layer 
tries to separate them. If the dimension of the Euclidean space in the last but one layer is 
bigger or equal to 
$N-1$, there  exists a softmax classifier to separate those clusters with probability 1. 
But if the dimension of the Euclidean space in the last but one layer is less than $N-1$, 
there may exist a cluster where the center is inside the convex closure of the other 
cluster centers. In this case, there is no softmax classifier that can separate 
this cluster from other clusters, because a linear function on a convex set
always take its maximal value at a vertex. (For a image, see Figure \ref{unseparable}. For more detail, see section   \ref{analysis}.) 
  


In order to solve the classification problems of $N$ classes  with growing $N$, 
either the the dimension in the last but one 
layer is fixed, which leads to that the performance is  bad,
or the dimension in the last but one 
layer grow with the growing of $N$, which leads to that the parameter number in the last two layers grow
hyper-linearly with the growing of $N$. 
The hyper-linear growth of the network size increases the training time and memory usage significantly, 
which limits many real applications that require huge number of class labels. 

This paper proposes a method so-called Label Mapping to solve this contradiction.
Our idea is to reduce a multi-class classification 
problem with huge number of classes to several multi-class classification 
problems with middle number of classes. Every multi-class classification 
problems with middle number of classes can be trained parallel. When we train them distributedly, the cost of storage
and computing in a single machine increase slowly with the increasing of the class number. Moreover, the communication 
between the machines is not needed.

A similar method to our method is Error-correcting output codes (ECOC), which is discussed in 
\cite{Dietterich_and_Bakiri}, \cite{Allwein_and_Schapire}, \cite{Passerini_Pontil_and_Frasconi} 
and etc..  It reduces a n-class classification problem to
several binary classification problems.  The 
ECOC typically applies binary classifier such as SVM, and therefore 
the binary error-correcting code is naturally used in this case. 

However, if we use a deep learning network as a 
base learner, it is not necessary to limit the code to be binary. In fact, there is a trade-off between
the class number of one base learner and the number of base learner used. According to information theory,
if we use  $p$ classes classifiers as basic classifiers to solve a classification problem of $N$-class,
we need at least $\lceil \log _p N \rceil $'s base learners. For example, if we need to solve a classifying problem of 1M's classes, 
and we use the binary classifier as base learners, we need at least 20 base learners. 
For some classical applications, for example, the CNN image 
classification, we need to build a CNN network for every binary classifier. It is huge cost for 
computation and memory resources. But if we combine different base learners with 1000 classes, we 
need only 2 base learners. 

In order to combine several multi-label base learners, the 
ECOC is not usable. Our Label Mapping (LM) method is very suitable for this purpose.
 
We discuss the design principles for LM, so-called  ``classes high separable'' and ``base learners independence''
. The principle ``classes high separable'' ensures that for any two different classes, 
there are as many as possible base learners 
are trained to separate them. The principle ``base learners independence'' ensures 
that the repeat part of the information learned by any two different base learners is as few as possible. 

Then we propose two classes of LM and prove that they 
conform with these principles.
   

As numeric experiments, we show the accuracies
of LM on three dataset, namely, the dataset Cifar-100, dataset CJK characters and the 
dataset ``Republic''. 
On all the datasets,
the accuracies of LM increase remarkably with the increasing of the number of base learner. 
When the class number is bigger than the  
dimension of the last but one layer of the 
 network (dataset CJK characters), 
the accuracy of LM is better than the one-hot encoding with standard softmax and negative sampling with same
number of parameters of network.  When the class number is much bigger than the dimension of the last but one layer of the 
 network (dataset ``Republic''),  the accuracy of LM is much better than the one-hot encoding with standard softmax and negative sampling with  bigger number of parameters of 
 network.
 
The base learners can be trained parallel, when we train them distributedly the cost of storage
and computing in a single machine increase slowly. (In fact, at most 
$O(\sqrt{N})$, where $N$ is the number of classes.) Moreover, the communication 
between the machines is not needed.

We compare LM with the classical method ECOC also, the accuracy of LM+DNN is much
greater than the accuracy of ECOC+DNN of the same and even bigger number of 
parameters.

%
 
This paper is organized as follows. In section 2, we give a literature review.
In section 3, we discuss the week point of the classifier of one-hot encoding with softmax.
In section 4, we give the formula definition of the LM, discuss the principle to design
LM and propose two classes of the LM and prove that the principles were satisfied. In section 
5, we give some numeric examples. 

There are some symbols used in this paper:

1). For a positive integral number $N$, $\mathbb{Z}/N\mathbb{Z}$ denote the set
$\{0, 1, \cdots, N-1\}$; 

2). for a power $q$ of a prime number,
$\mathbb{F}_q$ denote the finite field (Galois field) of $q$ elements;  

3). for a field $\mathbb{F}$, $\mathbb{F}[x]$ denote the polynomial ring on  
$\mathbb{F}$;

4). for a polynomial $g(x)$, $\deg g(x)$ denote the degree of $g(x)$.

\section{Literature review}

There are some researches about the multi-class classification problem  of huge number of classes using DNN,
for example, the hierarchical softmax  method \cite{hierarchical_softmax} and negative sampling 
 method \cite{negative_sampling}. These method can reduce the computational 
 complexity of train, but can not reduce the number of parameters. Their 
  performance is bad than  
 the standard one-hot plus softmax method also.

There is a method reduced a multi-class classification problems of big number of 
classes to several binary classification problem, i.e, the ECOC. 
T. G. Dietterich and G. Bakiri in \cite{Dietterich_and_Bakiri} introduced ECOC to combine several binary
classifiers to solve multi-class classification problems. In that 
paper, the design principles of ECOC ``Row separation'' and ``Column separation'' 
are proposed. In that paper, a decision tree C4.5 and a shallow neural network with sigmoid output are used 
as binary base learners. The ``exhaustive codes'' (equivalence to Hadamard code) for class numbers $3-7$, column 
selection from   exhaustive codes for class number $8-11$, and random hill 
climbing or BCH codes for class num bigger than 11 are used. For decoder,
it minimizes the L1 distance between the codewords and output probabilities.

In \cite{Allwein_and_Schapire}, Allwein and Schapire proposed to use symbols from $\{-1,0,1\}$ instead of $\{-1,1\}$
in encode. The output bits
which take value 0 in the encoded label does not appear in Loss. 
Using this modification, the three approaches, namely, 
one versus one, one versus others, and binary encode are collected into a common framework. 

In \cite{Escalera_Pujol_and_Radeva2}, Escalera, Pujol and Radeva discussed
the design principles of the modified binary ECOC which may take value in $\{-1,0, 1\}$, 
and gave some examples satisfying these principles.

In \cite{Passerini_Pontil_and_Frasconi}, Passerini, Pontil and Frasconi
disscused the decode method and the combining of ECOC with SVM with kernels.

In \cite{Langford_and_Beygelzimer}, Langford and Beygelzimer proposed a
reduction from cost-sensitive classification to binary classification based on a modification of ECOC.   


In \cite{Escalera_Pujol_and_Radeva}, \cite{Pujol_Radeva_and_Vitria}, 
\cite{Pujol_Escalera_and_Radeva},  \cite{Fa_Zheng_Hui_Xue_and_Xiaohong_Chen_Yunyun_Wang}, 
some applications dependent ECOC are proposed. The code-book is generated based on a discrimination tree. 

In \cite{Niloufar_Eghbali_and_Gholam_AliMontazer}, another class of application dependent ECOCs are proposed. It is constructed with considering the neighborhood of samples.

In \cite{AAAI}, ECOC is used to the representation learning.

In \cite{Zero-Shot}, ECOC is used to zero-shot action recognition.

In \cite{Ghani_KDD_report}, the ECOC is used to the text classification problem with a large number of categories.




Up to now, all the codes used in ECOC are binary, and all the basic classifiers 
used in ECOC are two-class classifiers.

\section{Analysis for classification using deep neural network}
\label{analysis}
Generally, A DNN classier
of $N$ classes can be treated as a series connection of a complex mapping in Euclidean space to the
last but one layer, and a softmax classifier softmax$(Ax+b)$ of $N$ classes in the last layer. The complex mapping can be 
interpreted as a clustering process to cluster data based on their class labels, and the last layer tries to separate them.
%
%
%
%
%
%
%
But the softmax classifier softmax$(Ax+b)$ can separate all the $N$ classes in 
the Euclidean space only if the centers of the clusters satisfy the convex property as following.

\begin{figure}
    \begin{minipage}[t]{0.49\linewidth}
  \includegraphics[scale=0.25]{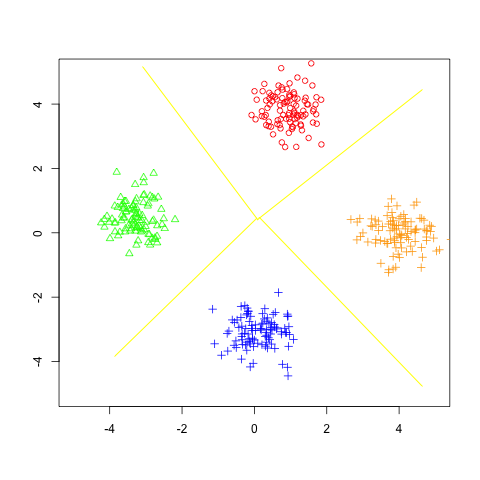}
\caption{The clusters are separable by softmax$(Ax+b)$.}
 \label{separable}
\end{minipage}
\begin{minipage}[t]{0.49\linewidth}
  \includegraphics[scale=0.25]{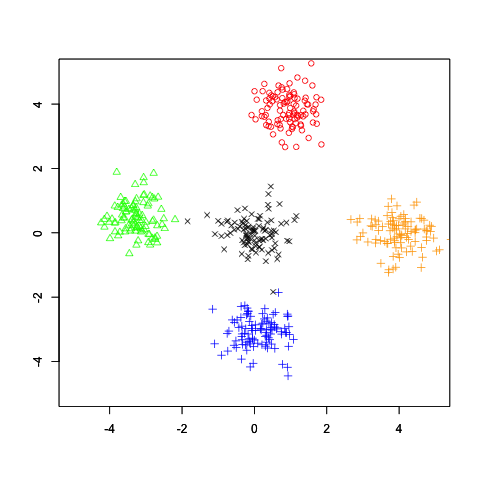}
\caption{The central cluster cannot be separated by any softmax$(Ax+b)$.}
\label{unseparable}
\end{minipage}
\end{figure}

\begin{defn}
  We call a set $X$ of $N$ points in an Euclidean space $V$  satisfies the convex 
  property if and only if the convex closure of $X$ has exact $N$ vertexes.
\end{defn}
 For example, the set of the centers of the 4 clusters in  Figure \ref{separable}  
 has the convex 
  property, but the set of the centers of the 5 clusters in  Figure \ref{unseparable}  
 has not the convex 
  property.

In other words, the softmax classifier softmax$(Ax+b)$ can separate all the $N$ clusters in 
the Euclidean space $V$ only if there are not any cluster, which center lie in the inner of the  convex closure of the centers
of other clusters. The reason is that, a linear function on a convex body can not take its max 
value in the inner of the body (Figure \ref{unseparable}).

If the dimension of the Euclidean space $V$ in the last but one layer is bigger or equal to $N-1$, 
the centers of the clusters satisfy the convex property with 
probability 1 (unless the $N$ centers lie in an affine subspace of dimension less than $N-1$ of 
$V$, which probability is $0$), and hence
there  exists a softmax classifier to separate those clusters with 
probability 1. 

But if the dimension of the Euclidean space in the last but one layer is less than $N-1$, 
the probability of that the centers of the clusters satisfy the convex property is less than 
1.
Moreover, when the dimension of the Euclidean space $V$ in the last but one layer is 
fixed, the probability of that  the centers of the clusters satisfy the convex property 
decrease with the increasing of $N$. If the class number $N$ is much bigger than 
the dimension of  $V$, the complex mapping in font layers in the network  
difficult to map the clusters such that the centers of the clusters satisfy the convex property. 
Hence there are not any softmax classifier can separate them.

\section{Label Mapping}

For a $N$-class classification problem,
we define a Label Mapping (LM) as a sequence of map 
\begin{displaymath} 
f_i: \mathbb{Z}/N\mathbb{Z} \longrightarrow \mathbb{Z}/N_i\mathbb{Z}, \quad i=1,2,...n, 
\end{displaymath} 
where each $f_i$ is called a ``site-position function'', and $n$ is called the ``length'' of the
Label Mapping.
If all the $N_i$ are equal to each other, we call it ``simplex LM''; otherwise 
we call it ``mixed LM''. 

\begin{figure}
   \begin{minipage}[t]{0.5\linewidth}
 \includegraphics[scale=0.35]{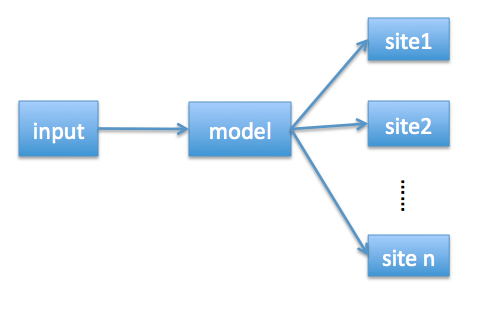}
\caption{One network with n outputs}
\label{joint_LM}
\end{minipage}
\begin{minipage}[t]{0.5\linewidth}
 \includegraphics[scale=0.35]{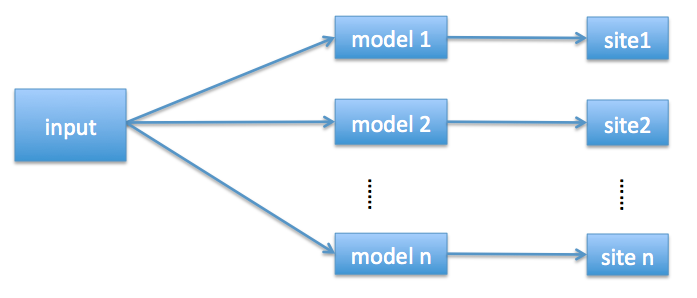}
\caption{n networks with n outputs}
\label{separable_LM}
\end{minipage}
\end{figure}

Generally, $N$ is a huge number, and $N_i$ are some numbers of middle size. We can
reduce a $N$-classes classification problem to $n$'s classification problems of 
middle size through a LM. Suppose the training dataset is $\{x_k, y_k\}$, where $x_k$ is 
feature and $y_k$ is label, there are two method  to use DNN plus LM. 
The one is to use one network with $n$ outputs (Figure \ref{joint_LM}). The other 
one is to use $n$ networks, every network is
trained as a base learner on
the dataset $\{x_k,f_i(y_k)\}$ for $i=1,2, \cdots n$ (Figure \ref{separable_LM}). 
Considering the convenience of distributed training, we use method in Figure \ref{separable_LM}.

A good LM should satisfy the follow properties:

\textbf{Classes high separable}. For two different labels $y, \tilde{y}$, there should be as 
many as possible site-position functions $f_i$ such that $f_i(y)\neq f_i(\tilde{y})$.

\textbf{Base learners independence}. When $y$ are selected randomly uniformly from $\mathbb{Z}/N\mathbb{Z}$,
the mutual information of $f_i(y)$ and $f_j(y)$ approximate to 0 for $i\neq j$.

The property ``classes high separable'' ensures that for any two different classes, 
there are as many as possible base learners are trained to separate them.
 The property ``base learners independence'' ensures that the common part 
 of the information learned by any two different base learners is as few as possible.

\textbf{Remark.}
There are some similarities between LM and ECOC:

A ECOC of length $n$ for $N$ classes can be regarded as a sequence of maps
\begin{displaymath} 
f_i: \mathbb{Z}/N\mathbb{Z} \longrightarrow \mathbb{Z}/2\mathbb{Z}, \quad i=1,2,...n, 
\end{displaymath} 
where each $f_i$ is called ``bit-position function''.

We can see that comparing to ECOC, our LM does not need that the reduced classification 
problems are two-class classification problems. It even does not need that
the reduced classification 
problems have the same number of classes.

In \cite{Dietterich_and_Bakiri}, two properties which
a good error-correcting output binary code for a multi-class problem should 
be satisfied are proposed: (\cite{Dietterich_and_Bakiri}, section 2.3)

\textbf{Row separation}. Each codeword should be well-separated in Hamming distance
from each of the other codewords.

\textbf{Column separation}. Each bit-position function should be uncorrelated with the
functions to be learned for the other bit positions.    

The property ``Row separation'' is similar to the property ``Classes high separable'' 
of LM, and the property ``Column separation'' is similar to the property ''Base learners 
independence''. \qed

 We give
2 classes of LM, one is mixed  LM, and the other is simplex LM, which satisfies the
properties.

\subsection{Mixed LM}

We introduce the famous Prime Number Theorem ( \cite{Riemann}, \cite{Prime_Number_Theorem}) firstly:
\begin{thm}(Prime Number Theorem)
  The density of prime number  approximate $n$ is $1/\log n$.  \quad $\blacksquare$
\end{thm}

For the original label's set $\mathbb{Z}/N\mathbb{Z}$, a small number k like 2, or 3, etc., and a small positive number
$\epsilon \in (0,1)$, select $n$'s  prime numbers
 $p_1, p_2, \cdots p_n$ in the domain $\left[N^\frac{1}{k}, N^\frac{1}{k-\epsilon}
\right)$. According to the Prime Number Theorem, there are about $\frac{ k (N^\frac{1}{k-\epsilon}-N^\frac{1}{k} )}{\log N}$ 
 prime numbers in this domain.

We define a LM as 
\begin{displaymath}
  \begin{array}{ccc}
    \mathbb{Z}/N\mathbb{Z} & \longrightarrow & \prod _{i=1}^n \mathbb{Z}/p_i\mathbb{Z} \\
    x & \mapsto & f_i(x) 
  \end{array}
\end{displaymath}
where $f_i(x)=x \mod p_i$. Then we have the following proposition:
\begin{thm}
  For any $x \neq y \in \mathbb{Z}/N\mathbb{Z}$, there is at most $k-1$'s $i$ s.t 
  $f_i(x)=f_i(y)$.
\end{thm}

\textbf{Proof.} Suppose there exist $k$'s different $i$ 
such, that $f_i(x)=f_i(y)$, we can suppose that
\begin{displaymath}
  f_i(x)=f_i(y) \quad \mbox{ for }i=1,2, \cdots k
\end{displaymath} 
Then we have $x \equiv y \mod p_i$ for all $i=1,2, \cdots, k$.

Because $\{p_i\}$ are prime numbers, we have $x \equiv y \mod \prod_{i=1} ^k p_i$. But we 
know $x, y \in \{0, 1, \cdots N-1 \}$, which in $\{0, 1, \cdots \prod_{i=1} 
^k p_i-1\} $, hence $x=y$.
$\blacksquare$

This theorem tells us that the mixed LM satisfies the ``Classes high separable'' property. 
Following, we prove that it satisfies the property ``Base learners independence''.

\begin{thm}
Let $x$ be uniformly randomly selected from $\mathbb{Z}/ N\mathbb{Z}$, we have that  
for any $i \neq j$,
  the mutual Information of $f_i(x)$ and $f_j(x)$ approximate 0.   
\label{Base_learners_independence}
\end{thm}

In order to prove this theorem, we give a lemma firstly.
\begin{lem}
  Let $x$ be uniformly randomly selected from $\mathbb{Z}/ N\mathbb{Z}$,  
$q$ is an positive integral number, $t:= \lceil \frac{N}{q} \rceil$. Then we have 
that the probability 
of $(x \mod q)$ at every point in $\mathbb{Z}/q\mathbb{Z}$ are $\frac{t}{N}$ or 
$\frac{t-1}{N}$.
\label{preimage}
\end{lem}

\textbf{Proof.} Because the pre-image of every point in $\mathbb{Z}/q\mathbb{Z}$ 
under the map 
\begin{displaymath}
  \begin{array}{rcl}
     \mathbb{Z}/N\mathbb{Z} & \longrightarrow & \mathbb{Z}/q\mathbb{Z} \\
     x & \mapsto & (x \mod q)
  \end{array}
\end{displaymath}
is a set of $t$ or $t-1$ elements. \qed

Now, we proof the theorem \ref{Base_learners_independence}:

\textbf{Proof of Theorem \ref{Base_learners_independence}:} Let $t_i:= \lceil \frac{N}{p_i} \rceil$ and  $s_{ij}= \lceil \frac{N}{p_ip_j} \rceil$ for every $i, j$. 
We have that the 
probabilities
of $f_i(x)$ at every point in $\mathbb{Z}/p_i\mathbb{Z}$ are $\frac{t_i}{N}$ or $\frac{t_i-1}{N}$ and
the  probabilities
of $(f_i(x), f_j(x))$ at every point in  $\mathbb{Z}/p_i\mathbb{Z} \times  \mathbb{Z}/p_j\mathbb{Z} \cong \mathbb{Z}/p_ip_j\mathbb{Z}$ 
are $\frac{s_{ij}}{N}$ or $\frac{s_{ij}-1}{N}$
by using the lemma \ref{preimage}.

We know that the mutual information of 
$y_i=f_i(x)$ and $y_j=f_j(x)$ is 
\begin{displaymath}
 I(Y_i;Y_j)=\sum_{(y_i,y_j)\in \mathbb{Z}/p\mathbb{Z} \times \mathbb{Z}/p\mathbb{Z}}  
 p_{i,j}(y_i, y_j)\log \frac{ p_{i,j}(y_i, y_j)}{p_i(y_i)p_j(y_j)}
\end{displaymath}

a.) When $k=2$, we have $N<p_ip_j$ and hence $s=1$ and $p_{i,j}(y_i, y_j)=\frac{1}{N}$ on $N$'s
point in $\mathbb{Z}/p\mathbb{Z} \times \mathbb{Z}/p\mathbb{Z}$ and $0$ on 
other points. Hence we have
\begin{displaymath}
  \begin{array}{rl}
 I(Y_i;Y_j) & \leq \sum_{(y_i,y_j)\in \mathbb{Z}/p\mathbb{Z} \times \mathbb{Z}/p\mathbb{Z}}  
 p_{i,j}(y_i, y_j)\log \frac{ p_{i,j}(y_i, y_j)}{\frac{(t_i-1)(t_j-1)}{N^2}} \\
 & =N \times \frac{1}{N} \log \frac{ 1/N}{\frac{(t_i-1)(t_j-1)}{N^2}} \\
& =\log N-\log(t_i-1) -\log (t_j-1)  \\
& < \log N -\log(\frac{N}{p_i}-1) -\log(\frac{N}{p_j}-1) \\
& \leq \log N - 2 \log( \frac{N^\frac{1}{2}}{N^{2-\epsilon} }-1) \\
& = -2\log(\frac{1}{N^{2-\epsilon}}-N^{-\frac{1}{2}}) \\
& \rightarrow 0 \quad \mbox{ as } N\rightarrow \infty
 \end{array}
\end{displaymath}

b.) When $k\geq 3$, we have $p_ip_j<N^\frac{2}{k-\epsilon}<N$, and

 \begin{displaymath}
   \begin{array}{rl}
     & I(Y_i;Y_j)     \\
     = & \sum_{(y_i,y_j)\in \mathbb{Z}/p_i\mathbb{Z} \times \mathbb{Z}/p_j\mathbb{Z}}  
 p_{i,j}(y_i, y_j)\log p_{i,j}(y_i, y_j)  \\
 & - \sum_{(y_i,y_j)\in \mathbb{Z}/p_i\mathbb{Z} \times \mathbb{Z}/p_j\mathbb{Z}}  
 p_{i,j}(y_i, y_j)(\log p_i(y_i)+ \log p_j(y_j))          \\
  =& \sum_{(y_i,y_j)\in \mathbb{Z}/p_i\mathbb{Z} \times \mathbb{Z}/p_j\mathbb{Z}}  
 p_{i,j}(y_i, y_j)\log p_{i,j}(y_i, y_j)  \\
 & - \sum_{y_i\in \mathbb{Z}/p_i\mathbb{Z}}  
 p_i(y_i)\log p_i(y_i)-  \sum_{y_j\in \mathbb{Z}/p_j\mathbb{Z}}  
 p_j(y_j)\log p_j(y_j)    \\
 \leq & p_i p_j \frac{s_{ij}}{N} \log(\frac{s_{ij}}{N})  -p_i \frac{t_i-1}{N}\log \frac{t_i-1}{N} 
 -p_j \frac{t_j-1}{N}\log \frac{t_j-1}{N}   
   \end{array}
 \end{displaymath}
 Because

 \begin{displaymath}
   \begin{array}{rcl}
     (s_{ij}-1)p_i p_j & <N \leq & s_{ij}p_ip_j \\
     (t_i-1)p_i   & <N \leq & t_ip_i \\
     (t_j-1)p_j   & <N \leq & t_j p_j
   \end{array}
 \end{displaymath}

We have 
 \begin{displaymath}
   \begin{array}{rl}
     & I(Y_i;Y_j)     \\
    < & (1+\frac{p_ip_j}{N}) \log (\frac{1}{p_ip_j}+\frac{1}{N}) \\
    &   -(1-\frac{p_i}{N}) 
    \log (\frac{1}{p_i}-\frac{1}{N})  -(1-\frac{p_j}{N}) 
    \log (\frac{1}{p_j}-\frac{1}{N}) \\
  = & \log \frac{\frac{1}{p_ip_j}+\frac{1}{N}}{(\frac{1}{p_i}-\frac{1}{N})(\frac{1}{p_j}-\frac{1}{N})}
  +\frac{p_ip_j}{N} \log (\frac{1}{p_ip_j}+\frac{1}{N}) \\
&  +\frac{p_i}{N}  \log (\frac{1}{p_i}-\frac{1}{N})
  +\frac{p_j}{N} \log (\frac{1}{p_j}-\frac{1}{N})  \\
 \leq & 
 \log(1+\frac{p_ip_j}{N})-\log(1-(\frac{1}{p_i}+\frac{1}{p_j})\frac{p_ip_j}{N}+\frac{1}{N^2}) 
 \\
 \leq & 
 \log(1+\frac{p_ip_j}{N})-\log(1-(\frac{1}{p_i}+\frac{1}{p_j})\frac{p_ip_j}{N}) 
 \\
 = & O(\frac{p_ip_j}{N})+O((\frac{1}{p_i}+\frac{1}{p_j})\frac{p_ip_j}{N}) \\
 = & O(\frac{p_ip_j}{N})\\
 = & O(N^{\frac{2}{k-\epsilon}-1}) \\
 = & O(N^\frac{2+\epsilon-k}{k-\epsilon}) 
  \rightarrow 0 \quad \mbox{as} \quad  N\rightarrow \infty     
   \end{array}
 \end{displaymath}

$\blacksquare$

This theorem tells us that, the mixed LM  satisfies the property ``Base learners independence''.

\subsection{Simplex LM}

\subsubsection{p-adic representation}
For any prime number $p$, we can represent any non-negative integral number $x$ less than $p^k$ as the 
unique form
$x=x_0+x_1p+ \cdots + x_{k-1} p^{k-1} \quad (x_i \in \mathbb{Z}/p\mathbb{Z})$, which gives a bijection
\begin{displaymath}
  \mathbb{Z}/p^k\mathbb{Z} \longrightarrow \mathbb{F}_p ^k
\end{displaymath}

For the classification problem of $N$-classes and any small positive integral number $k$ 
(for example, k=2, 3), let $\epsilon$ in (0,1) be a small positive real number, and 
 take the a prime number $p$ in the domain $[N^{\frac{1}{k}}, N^{\frac{1}{k-\epsilon}}]$. 
(By The Prime Number Theorem \cite{Prime_Number_Theorem}\cite{Riemann}, the number of prime number in the domain 
$[N^{\frac{1}{k}}, N^{\frac{1}{k-\epsilon}}]$ is about 
$k(N^{\frac{1}{k-\epsilon}}-N^{\frac{1}{k}})/\log N$
), and get a injection
\begin{displaymath}
  \mathbb{Z}/N\mathbb{Z} \longrightarrow \mathbb{Z}/p^k\mathbb{Z} \longrightarrow \mathbb{F}_p ^k
\end{displaymath}  
by p-adic representation.
Then we can combine this map with any injection $\mathbb{F}_p ^k \longrightarrow \mathbb{F}_p^n$ 
to get $p$-ary simplex LM.

\subsubsection{Singleton bound and MDS code}
In coding theory, the Singleton bound, named after Richard Colome Singleton 
\cite{Singleton}, is a relatively crude upper bound on the size of an arbitrary 
q-ary code $C$ with block length $n$, size 
$M$ and minimum distance $d$.

The minimum distance of a set C of codewords of length $n$ is defined as
\begin{displaymath}
d=\min _{\{x,y\in C:x\neq y\}}d(x,y) 
\end{displaymath}
where $d(x,y)$ is the Hamming distance between $x$ and $y$. 
The expression $A_{q}(n,d)$ represents the maximum number of possible 
codewords in a q-ary block code of length $n$ and minimum distance $d$.
Then the Singleton bound states that

\begin{thm}(Richard Collom Singleton) 
$A_{q}(n,d)\leq q^{n-d+1}$.  \quad $\blacksquare$
\end{thm}

The code achieving Singleton bound is called MDS 
(maximal distinct separate) code. 

It is easy to see that, for a fixed 
original ID's number $N$, code length $n$, MDS codes are the codes which 
most satisfies the property ``Class high separable''. Fortunately, for big prime number or power over prime number $q$, there are some
nontrivial MDS codes found, for example the Reed-Solomon code\cite{Reed_and_Solomon}.

\begin{thm}(Reed and Solomon)
 For $n$'s different elements $x_1, x_2, \cdots, x_n$ in $\mathbb{F}_q$, the code defined by the composite of the map 
 \begin{displaymath}
   \begin{array}{rcl}
     \mathbb{F}^k_q & \longrightarrow & \mathbb{F}_q[x]_{deg<k}  \\
     (a_0, \cdots, a_{k-1} ) & \mapsto & a_0+a_1x+ \cdots a_{k-1} x^{k-1} 
   \end{array}
 \end{displaymath} 
 and the map
  \begin{displaymath}
   \begin{array}{rcl}
     \mathbb{F}_q[x]_{deg<k}  & \longrightarrow & \mathbb{F}_q^n  \\
     f(x) & \mapsto & (f(x_1), \cdots f(x_n) ) 
   \end{array}
 \end{displaymath} 
 is a MDS code. \quad $\blacksquare$
  \end{thm}

In this paper, we use only the Reed-Solomon code with q=p be a prime number.

\textbf{Remark}. In the case of ECOC, the property  similar to ``Class high separable'' 
is ''Row separation''. If there exists a nontrivial binary MDS code, it will be the 
code most satisfies ``Row separation'' also.  
But unfortunately, it has not 
find any nontrivial binary MDS code yet up to now. In fact, for some situation,
the fact that there are not any nontrivial binary MDS code is proved.  
(\cite{Guerrini_and_Sala} and Proposition 9.2 on p. 212 in \cite{Vermani} ). This
is an advantage of simplex LM better than ECOC also.

\subsubsection{Separability and independency}
We can combine the $p$-adic representation map with a Reed-Solomon encoder 
over field $\mathbb{F}_p$ to get a simplex LM for any prime number $p$.
The above theorem ensures that, this code satisfies the 
property ``Classes high separable''. We will prove that, it satisfies the property ``Base learners independence'' also.



\begin{thm}
  \label{simplex_LM_Thm}
  If $u$ is a random variable with uniform distribution on 
$\mathbb{Z}/N\mathbb{Z}$, $y_i$ and $y_j$ are the i-site value and j-site value 
($i \neq j$) 
of the codeword of $u$ under the simplex LM described above, then the mutual information of 
$y_i$ and $y_j$ approach to $0$ when $N$ grows up. 
\end{thm}

The proof of this theorem is similar to the proof of the theorem 
\ref{Base_learners_independence},
we omit it  due to space limitations.

$\blacksquare$

\subsection{Decode Algorithm}
Suppose we used the LM
\begin{displaymath} 
f_i: \mathbb{Z}/N\mathbb{Z} \longrightarrow \mathbb{Z}/N_i\mathbb{Z}, \quad i=1,2,...n, 
\end{displaymath} 
to reduce a classification problem of class number $N$ to the classification problems of 
class number $N_i$'s, and trained $n$ base learner for every $f_i$, the output 
of every base learner $i$ is a distribution $P_i$ 
on $\mathbb{Z}/N_i\mathbb{Z}$. Now, for a input feature data,
how we collect the output $\{P_i : i=1,2,\cdots, n\}$ of every
base learner to get the predict label?
    
In this paper, we search the $a\in \mathbb{Z}/N\mathbb{Z}$ such that $\sum_i \log P_i(f_i(a))$ 
is maximal, and let such $a$ be the decoded label. 

In fact, $\sum_i \log P_i(f_i(a))=
-\sum_i KL(f_{i\star} \delta(x-a) ||  P_i)$
,  where $\delta(x-a)$ is the Delta distribution at $a\in \mathbb{Z}/N\mathbb{Z}$, and $f_{i\star} \delta(x-a)$ is 
the marginal distribution of $\delta(x-a)$ induced by $f_i$. This decode 
algorithm is that find the Delta distribution on $\mathbb{Z}/N\mathbb{Z}$ such 
that the marginal distribution on every $\mathbb{Z}/N_i\mathbb{Z}$ included from it is as closed to 
$P_i$ as possiple.

\section{Numeric experiments}

We give performance of LM on three dataset, namely, the dataset \textbf{Cifar-100} 
\cite{Cifar-100}
, the dataset \textbf{CJK characters} and the dataset \textbf{Republic}.
The CIFAR-100 dataset consists of 60000 32x32 color images in 100 classes, with  500 training images 
and 100 testing images per class. The dataset  \textbf{CJK characters} 
is the grey-level image of size 139x139 of 20901 CJK characters (0x4e00 $\sim$ 0x9fa5)
in 8 fonts. The dataset \textbf{Republic} is a text with 118684 words 
and 7409 unique words in the vocabulary.

We use a simple CNN network, which dimension in the last but one layer  is 128, 
with a one-hot encoding as the baseline for the cifar-100 dataset.

We use an inception V3 network \cite{inception_v3}, which dimension in the last but one layer  is 
2048, with a one-hot encoding as the baseline for the CJK characters
dataset.

We use a RNN network which dimension in the last but one layer  is 100, with a 
one-hot encoding as the baseline for the dataset ``Republic''.

We will see that the accuracy of LM increases with the increasing of its length 
on all the datasets.
But the accuracy of LM on (cifar-100, simple CNN) is difficult 
to be higher than the one-hot, the accuracy of LM on (CJK character, inception V3) 
better than one-hot with  the almost same number of parameters, and the accuracy of LM on
(Republic, RNN) is much better than one-hot with more number of parameters.

Why there is a such big difference of accuracy in three situations?
Because the dimension of the last but one layer of the simple CNN
is bigger than the class number of cifar-100 dataset, and hence the one-hot encoding can 
brought into full play the power of simple CNN. But the dimension of the last but one layer of
inception V3 is less than the class number of dataset CJK character, which 
causes that the one-hot encoding can not  bring the power of 
inception V3 into full play. Moreover, the dimension of the last but one layer of
RNN is much less than the class number of dataset Republic, which 
causes that the one-hot encoding absolutely can not  bring the power of 
RNN into full play  (According to the relation of the dimension of the last but one 
layer and the class number discussed in section \ref{analysis}.)
\subsection{On a dataset of small class number}

We use a simple CNN network on the dataset Cifar-100. 
The simple CNN network includes 3 convolution layers and a full-connected layer. The sizes of convolution kernels are $3\times 3$, and the weigth of the three convolution layers
are 32, 64, 128 respectively. After each convolution layer, a $2 \times 2$
average poling layer is applied. After the 3rd pooling layer and the full-connection layer, 
we use dropout layer of probability 0.25. The network structure is like in Figure 
\ref{simple_cnn}.


Note the dimension 128 of the last but one layer is greater than the class 
number 100, hence the one-hot encoding can  bring the power of 
simple CNN into full play. In this experiment the LM is difficult to surpass the one-hot, but 
we can see the accuracy increases with the increasing of length of LM. We will see
the accuracy of LM is greater than the accuracy of ECOC also.

\subsubsection{The performance of simplex LM of different length}

We use the simplex LM defined above with $p=11, k=2$ and $n=3,4,5,6$.
The simplex LM can be writn as
\begin{displaymath} 
f_i: \mathbb{Z}/N\mathbb{Z} \longrightarrow \mathbb{Z}/N_i\mathbb{Z}, \quad i=1,2, \cdots n 
\end{displaymath}
where $N=100$, and $f_i(x)=((x \mod p)+floor(x/p)i) \mod p$. 

 We train 
the networks with batch size=128 and 390 batch per epoch.
The Figure \ref{simplex_LM_diff_len} shows the accuracy of these simplex LMs with the single CNN on 
dataset cifar-100:

\begin{figure}
  \begin{minipage}[t]{1.0\linewidth}
      \includegraphics[scale=0.2]{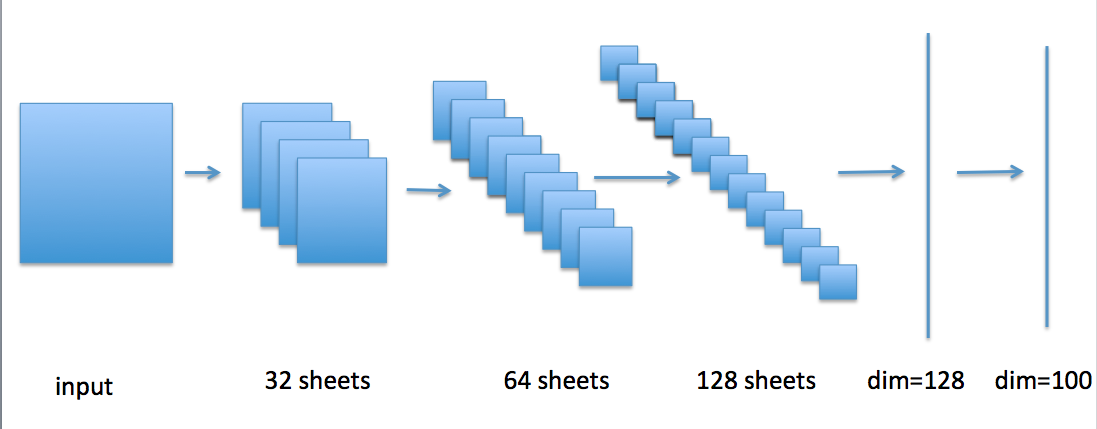}
\caption{The network structure of simple CNN}
\label{simple_cnn}
  \end{minipage}
\begin{minipage}[t]{1.0\linewidth}
\includegraphics[scale=0.5]{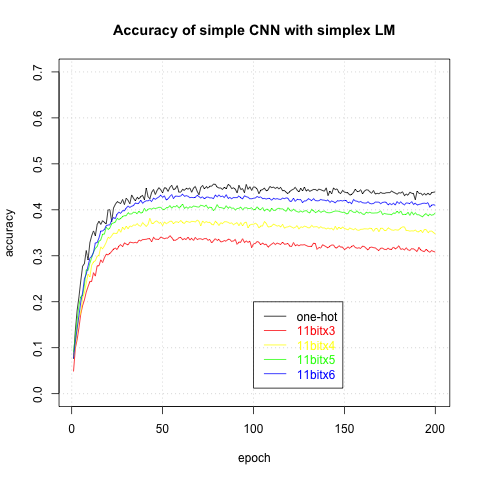}
\caption{The accuracy of simplex LMs of n=3,4,5,6 and one-hot}
\label{simplex_LM_diff_len}
\end{minipage}
\end{figure}

%

In Figure \ref{simplex_LM_diff_len}, the horizontal axis is the traning
epoch, and the vertical axis is the validation accuracy. 
The five curves, which colors are red, yellow, green, blue and black respectively, are
the epoch-accuracy curves of the simple CNN with simplex LM defined above with $p=11, k=2$ and $n=3,4,5,6$ 
and the one-hot encoding respectively. 
    
We can see, the accuracy of these networks with simplex LM and one-hot encoding increase in the 
first 50$\sim$80 epoch, and then a little of overfitting occur. The accuracy of simple 
CNN with simplex LM increases with the increasing of length of the LM, it
approximates the
accuracy of one-hot as $n$ increase, but it is difficult to surpass the one-hot. 
The reason is that the dimension of the last but one layer of the simple CNN
is bigger than the class number of cifar-100 dataset, and hence the one-hot encoding can 
bring the power of simple CNN into full play.

\subsubsection{The performance of mixed LM of different length}

We use the mixed LMs defined above with $k=2$ and $\{p_i\} \subset \{11,13, 17,19, 23\}$.
 The mixed LM can be writen as
\begin{displaymath} 
f_i: \mathbb{Z}/N\mathbb{Z} \longrightarrow \mathbb{Z}/p_i\mathbb{Z}, \quad i=1,2, \cdots n 
\end{displaymath}
where $N=100$, and $f_i(x)=x \mod p_i$.

The Figure \ref{mixed_LM_diff_len} shows the accuracy of these mixed LMs with the simple CNN on 
dataset cifar-100:


In Figure \ref{mixed_LM_diff_len}, the horizontal axis is the traning
epoch, and the vertical axis is the validation accuracy. 
The five curves, which colors are red, yellow, green, blue and black respectively, are
the epoch-accuracy curves of the simple CNN with simplex LM defined above with $\{p_i\}$
are $\{11, 13\}$, $\{11, 13, 17\}$, $\{11, 13, 17, 19\}$, $\{11, 13, 17, 19, 23\}$ 
and the one-hot encoding respectively. 
    
We can see, the accuracy of these networks with mixed LM and one-hot encoding increase in the 
first 50$\sim$80 epoch, and then a little of overfitting occur. The accuracy of simple 
CNN with mixed LM increases with the increasing of length of the LM, it
approximates the
accuracy of one-hot as $n$ increase, but it is difficult to surpass the one-hot. 
The reason is that the dimension of the last but one layer of the simple CNN
is bigger than the class number of cifar-100 dataset, and hence the one-hot encoding can 
brought into full play the power of simple CNN.

%

\subsubsection{Compare LM with ECOC}

On this subsection, we compare the performance of LM and ECOC.
We show the accuracy of the following LMs (ECOC is a special case of LM) with the simple CNN network
on the dataset cifar-100:

a. A ECOC of n=7

b. A simplex LM of p=11 and n=3.

c. A mixed LM of $\{p_i\}=\{11,13,17\}$. 

For the ECOC of $n=7$, we encode a label $y \in \{0, 1, \cdots, 99 \}$ to its 
binary representation of length 7.

The simplex LM can be written as
\begin{displaymath} 
f_i: \mathbb{Z}/N\mathbb{Z} \longrightarrow \mathbb{Z}/N_i\mathbb{Z}, \quad i=1,2, \cdots n 
\end{displaymath}
where $N=100$, and $f_i(x)=((x \mod p)+floor(x/p)i) \mod p$.

 The mixed LM can be writn as
\begin{displaymath} 
f_i: \mathbb{Z}/N\mathbb{Z} \longrightarrow \mathbb{Z}/p_i\mathbb{Z}, \quad i=1,2, \cdots n 
\end{displaymath}
where $N=100$, and $f_i(x)=x \mod p_i$.

\begin{figure}
  \begin{minipage}[t]{0.49\linewidth}
\includegraphics[scale=0.25]{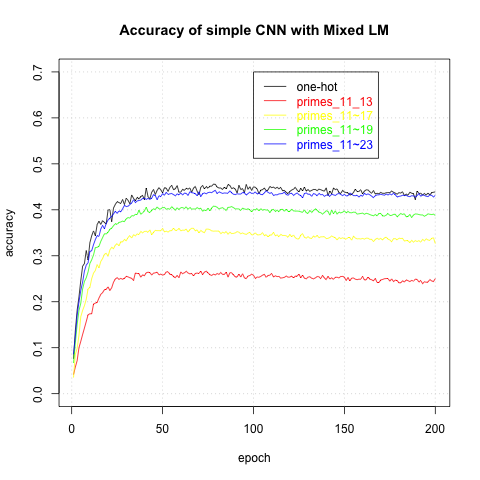}
\caption{The accuracy of mixed LMs of n=2,3,4,5 and one-hot}
\label{mixed_LM_diff_len}
  \end{minipage}
\begin{minipage}[t]{0.49\linewidth} 
   \includegraphics[scale=0.25]{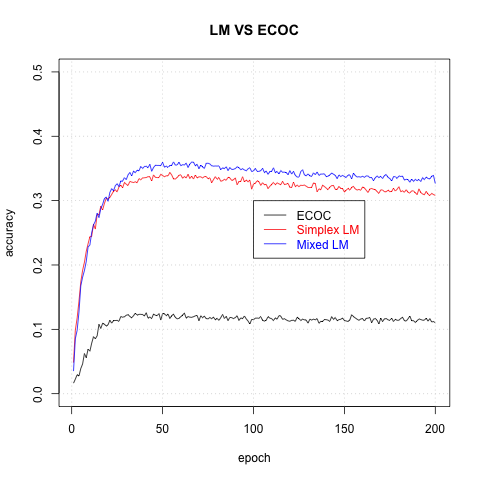}
\caption{Compare LMs with ECOC}
\label{LM_ECOC_7}
\end{minipage}
\end{figure}

The number of parameters of the three method are 1112622, 480321, 481353.
The accuracies of the three method is like in Figure \ref{LM_ECOC_7}. 
We can see, the accuracies of LMs are better than the ECOC even when the number 
of parameters of LM is much less than one of ECOC.

\subsection{On a dataset ``CJK Characters'' of big class number}

\begin{figure}
  \includegraphics[scale=0.5]{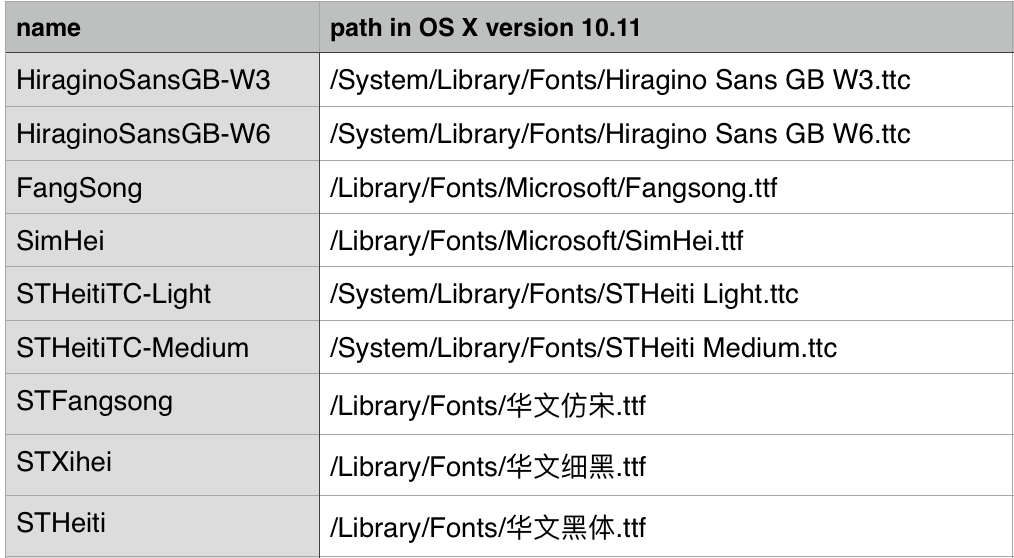}
\caption{The fonts and their paths of fonts in OS X version 10.11}
\label{fonts}
\end{figure}

We use the Inception V3 network and LM on the  dataset ``CJK characters''. CJK is a collective term for the Chinese, Japanese, and Korean languages, all of which use 
Chinese characters and derivatives (collectively, CJK characters) in their writing systems. 
The data set ``CJK characters'' is the grey-level image of size 139x139 of 20901 CJK characters (0x4e00 $\sim$ 0x9fa5)
in 8 fonts
, these fonts are in the paths in Figure \ref{fonts} of a MacBook Pro with OS X version 10.11.


 We use 7
fonts as the train set, and other one font as the test set.  We use inception v3 
network as base learner, and train the networks using batch size=128 and 100 batch per an epoch.

Noted the dimension 2048 of the last but one layer of Inception V3 
is much less than the class 
number 20901, hence the one-hot encoding can not bring the power of 
simple CNN into full play. In this experiment  
we can see the accuracy increases with the increasing of length of LM, and
the LM is easy to surpass the one-hot. We will see
the accuracy of LM is greater than the accuracy of ECOC also.

This section is divided into 3 parts. The part 1, 2 shows the performance of simplex LM and
mixed LM  respectively. 
Their accuracies 
increase with the increasing of length, and surpass the one-hot when length is greater or equal to 3. 
The part 3 compares the performance of LMs with the ECOC. 

\subsubsection{The performance of simplex LM of different length}

In this part, we shows how the accuracy of simplex LM increases with the increasing of length
of LM. We also show that when the length increases, the performance of LM is 
better than the network with one-hot encoding.

We use the simplex LM with $p=181$ and $k=2$. The simplex LM can be writn as

\begin{displaymath} 
f_i: \mathbb{Z}/N\mathbb{Z} \longrightarrow \mathbb{Z}/N_i\mathbb{Z}, \quad i=1,2, \cdots n 
\end{displaymath}
where $N=21901$, and $f_i(x)=((x \mod p)+floor(x/p)i) \mod p$. The accuracies 
of theses simplex LMs are like in Table
\ref{simplex_peroformance}.

In the table \ref{simplex_peroformance}, the accuracies in epoch 20, 40, 60, 80 
are showed. The column ``epoch'' means the epoch number in training, the column ``n sites''
is the accuracy of the inception V3 networks with simplex LM of n sites for $n=2,3, \cdots 6$, and the column ``one-hot''
is the accuracy of the inception V3 with one-hot encoding.

We see that, the accuracies of LM of all the site numbers increase with the increasing of training epoch.
The accuracy of LM increase with the increasing of sites number also. 
When the sites number is 
equal to 3, the parameters number of LM is approximate to the parameters number of one-hot, but the accuracy of LM
is greater than the one-hot with softmax or negative sampling.

\textbf{Remark}. In the one-hot network in the table \ref{simplex_peroformance},
the last layer has dimension 20901, but the last but one layer has dimension only 
2048. If we set the dimension of the last but one layer to 20900, the 
performance may be better, but our GPU does not have such huge memory.

\textbf{Remark}. In the term ``one-hot with negative sampling''  in the table 
\ref{simplex_peroformance}, the using negative sampling ratio is 10:1.
\begin{table}[!hbp]
  \tiny
\begin{tabular}{|c|c|c|c|c|c|c|c|}
\hline
\hline
ep.  & 2 sites & 3 sites & 4 sites & 5 sites & 6 sites  &  one-hot & one-hot with\\
      &&&&&&  with   &  negative        \\
      &&&&&&  softmax  &  sampling  \\
\hline
20	&	0.0118 & 0.0318	 &	0.0604 & 0.0585 & 0.0640	& 	  	 0.0325 & 0.0004\\
 40 & 0.6657 & 0.9373 & 0.9812 & 0.9865 &	0.9878 & 0.5152 & 0.0007\\
60	& 0.8172 & 0.9840 & 0.9943 & 0.9964 & 0.9968 & 
0.9399 & 0.0019\\
80 & 0.8684 & 0.9920 & 0.9978 & 0.9984 & 0.9988 & 		0.9854 & 0.0031\\	
\hline
\hline
param.  num.  &  $2.21 $ &  $2.21 $ & $2.21 $ & $2.21 $ & $2.21 $  & 6.46 & 6.46\\ 
($10^7$) & $\times 2$ & $\times 3$ & $\times 4$ & $\times 5$ & $\times 6$  & &\\
\hline
\hline
\end{tabular}
  \caption{Performance of Simplex Label Mappings}
\label{simplex_peroformance}
\end{table}

\subsubsection{The performance of mixed LM of different length}
%
%
%
%
%
%

In this part, we show how the accuracy of mixed LM increases with the increasing of length
of LM. We also show that when the length increases, the performance of LM is 
better than the network with one-hot encoding.

We use the mixed Label Mappings with primes in \{149, 151, 157, 163, 167, 173,
179\}.  The mixed LM can writn as
\begin{displaymath} 
f_i: \mathbb{Z}/N\mathbb{Z} \longrightarrow \mathbb{Z}/N_i\mathbb{Z}, \quad i=1,2, \cdots n 
\end{displaymath}
where $N=21901$, and $f_i(x)=x \mod p_i$, $p_1=149, p_2=151, \cdots$.

The accuracies are like in following table \ref{mixed_peroformance}. In this table, the accuracies in epoch 20, 40, 60, 80 
are showed. The column ``epoch'' means the epoch number in training, the column ``n sites''
is the accuracy of the inception V3 networks with simplex LM of n sites for $n=2,3, \cdots 7$, and the column ``one-hot''
is the accuracy of the inception V3 with one-hot encoding.

We see that, the accuracies of mixed LM of all the site numbers increase with the increasing of training epoch.
The accuracy of mixed LM increases with the increasing of sites number also. 
When the sites number is 
equal to 3, the parameters number of LM is approximate to the parameters number of one-hot, but the accuracy of LM
is greater than the one-hot.  

\textbf{Remark}. In the one-hot network in the table \ref{mixed_peroformance},
the last layer has dimension 20901, but the last but one layer has dimension only 
2048. If we set the dimension of the last but one layer to 20900, the 
performance may be better, but our GPU does not have such huge memory.

\textbf{Remark}. In the term ``one-hot with negative sampling''  in the table 
\ref{mixed_peroformance}, the using negative sampling ratio is 10:1.

\begin{table}[!hbp]
   \tiny
\begin{tabular}{|c|c|c|c|c|c|c|c|c|}
\hline
\hline
ep.  & 2 sites & 3 sites & 4 sites & 5 sites & 6 sites & 7 sites &  one-hot with & one-hot with\\
&&&&&&& softmax & negative sampling \\
\hline
20	&	0.0081 & 	0.0101 &	 0.0100 &	0.0309 & 	0.0585 & 	0.0926 & 	 0.0325 & 0.0004\\
 40 & 0.6130 & 	0.8100 &	 0.8707 &	0.9656 & 	0.9851 & 	0.9903 & 	 0.5152 & 0.0007\\
60	& 0.7629 &	0.9765 &	 0.9925 &	0.9957 &	0.9967 & 	0.9974 & 	 
0.9399 & 0.0019\\
80 & 0.8757 &	0.9912 & 	0.9971 & 	0.9980 & 	0.9982 & 	0.9987 & 		0.9854 & 0.0031\\	
\hline
\hline
param. num. &  $2.21 $ &  $2.21 $ & $2.21 $ & $2.21 $ & $2.21 $ & $2.21 $ & 6.46 & 6.46\\ 
 ($10^7$) & $\times 2$ & $\times 3$ &$\times 4$& $\times 5$&$\times 6$&$\times 7$&  &\\
\hline
\hline
\end{tabular}
  \caption{Performance of Mixed Label Mappings}
\label{mixed_peroformance}
\end{table}

\subsubsection{Compare LM with ECOC}

We show the arccuracies of the following ensemble methods with the inception V3 network
on the CJK dataset:

a. A 15 bits ECOC corresponding to the binary representation of label $0 \sim 20900$

b. A 2 sites simplex LM of p=181 and n=2

c. A 2 sites mixed LM of p in \{149, 151\}

The three settings are the minimal setting for the three methods respectively, it 
means, if we reduce any bit of the encoding or any site of the label mapping, the 
encoding or the label mapping will be not injection. The accuracies are in Table 
\ref{ECOC_LM}:

\begin{table}[!hbp]
 \tiny
\begin{tabular}{|c|c|c|c|}
\hline
\hline
ep.  & ECOC of 15 bit & simplex LM of 2 sites &  mixed LM of 2 sites \\
\hline
20	& 0.0069 &  0.0118 &	0.0081   \\
 40 & 0.0795 & 0.6657 & 0.6130 	 \\
60	& 0.3660 &  0.8172 &  0.7629  \\
80 & 0.5740 &  0.8684 &       0.8757  \\
\hline
\hline
param. num. ($10^7$) & $2.18 \times 15$ & $2.21 \times 2$ &  $2.21 \times 2$ \\ 
\hline
\hline
\end{tabular}
  \caption{Comparing of ECOC and Label Mappings}
\label{ECOC_LM}
\end{table}

We can see, even when the base learner number 2 of LM is much less than the base learner number 15 of ECOC, 
and the parameters number of LM is much less than the parameters number of ECOC,
the performance of LM is better than the ECOC. 

\subsection{On the dataset  ``Republic''}

The Republic (\cite{Republic},\cite{Republic_text}) is a Socratic
 dialogue, written by Plato around 380 BC, concerning justice, the 
 order and character of the just, city-state, and the just man.

 We use the following produce firstly:
 
a). Replace `-' with a white space.

b). Split words based on white space.

c). Remove all punctuation from words.

d). Remove all words that are not alphabetic to remove standalone punctuation tokens.

e). Normalize all words to lowercase.

After the produce, there are 118684 words in the produced text, and 7409 unique words in the 
vocabulary.

We construct  a  network which use the 50 previous words as input and predict the current word. 
Because both the input and output are categorical with big number of classes, we use the LM method not only for output, but for input also.

 In fact, for a LM 
 \begin{displaymath} 
f: \mathbb{Z}/N\mathbb{Z} \longrightarrow  \prod _{i=1}^n \mathbb{Z}/N_i\mathbb{Z}
\end{displaymath} 
  We get a sparse encode method
 \begin{displaymath} 
f: \mathbb{Z}/N\mathbb{Z} \longrightarrow  \prod _{i=1}^n \mathbb{Z}/N_i\mathbb{Z} \longrightarrow 
\prod_{i=1}^n  \mathbb{F}_2^{N_i} \cong  \mathbb{F}_2 ^{\sum N_i}
\end{displaymath} 
  induced by this LM naturally. The  middle map is defined by $(x_i)_i \mapsto \mbox{ (The one-hot 
  encoding of }
  x_i \mbox{ of length } N_i)_i$. Use this map, we can get a $n$-hot code of length $\sum_i N_i$ for 
  every label in $\mathbb{Z}/N\mathbb{Z}$, which can be used as input encoding.

\begin{figure}
 \includegraphics[scale=0.2]{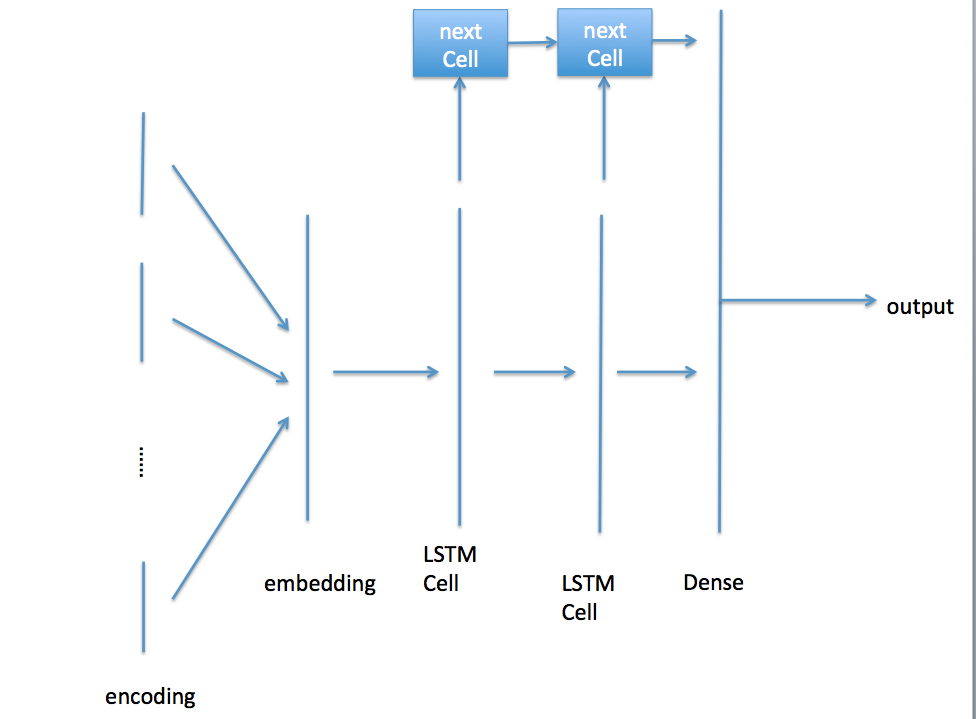}
\caption{The network structure}
\label{input_output_encode_lstm}
\end{figure}

  The network include an input encoding layer of dimension $\sum_i N_i$ , an embedding layer of 
  dimension 150, two LSTM layer of dimension 100, a dense layer of dimension 100, 
  and a dense output layer.
  After every  output layers of dimension $N_i$, a softmax is used. The 
  structure of the network is like in Figure \ref{input_output_encode_lstm}. 
  In Figure  \ref{input_output_encode_lstm} we draw only one encoding unit and 
  one embedding unit, in fact there are encoding unit and 
   embedding unit before every LSTM cell in first LSTM layer, but the weight of 
   the encoding units and  embedding units are same respectively.
  
%

The performance is like following table \ref{nlp_joint}, where the mixed LM of 2 sites use 
$\{N_i\}=\{107, 109\}$, the mixed LM of 4 sites use $\{N_i\}=\{107, 109, 113, 127\}$,
the mixed LM of 6 sites use $\{N_i\}=\{107, 109, 113, 127, 131, 137\}$. The 
simplex LM use the prime number $127$.

 \begin{table}[!hbp]
   \tiny
\begin{tabular}{|c|c|c|c|c|c|c|c|}
\hline
\hline
input & output & par. num. & 2 & 4 &  6 & 8  &  10 \\
\hline
one-hot & one-hot &  2.9E6 & 0.0946 & 0.1265 & 0.1379 & 0.1471 & 0.1540 \\
& softmax & &&&&& \\
\hline
one-hot & one-hot &  2.9E6 & 0.048 & 0.058 & 0.058 & 0.058 & 0.058 \\
& negative sampling &&&&&& \\
\hline
724 bit cut-off & 724 bit cut-off & 3.7E5 & 0.0589 & 0.1076 & 0.1245 & 0.1272 & 0.1321 \\
 \hline
  mix. LM of 6 sites &   mix. LM of 2 sites & 6.2E5 & 0.1331 & 0.1402 & 0.1366 & 0.1358 & 0.1189\\
 \hline
  mix. LM of 6 sites &   mix. LM of 4 sites & 1.2E6 & 0.1609 & 0.1722 & 0.1795 & 0.1836 & 0.1845 \\ 
 \hline
 mix. LM of 6 sites &   mix. LM of 6 sites & 1.9E6 & 0.1590 & 0.1731 & 0.1812 & 0.1849 & 0.1865 \\ 
\hline
sim. LM of 6 sites & sim. LM of 2 sites &  6.3E5 & 0.1453 & 0.1505 & 0.1531 & 0.1522 & 
0.1444 \\
\hline
sim. LM of 6 sites & sim. LM of 4 sites & 1.3E6 & 0.1586 & 0.1685 & 0.1759 & 0.1805 
& 0.1832 \\
\hline
sim. LM of 6 sites & sim. LM of 6 sites & 1.9E6 & 0.1575 & 0.1694 & 0.1776 & 0.1814 
& 0.1851 \\
\hline
\hline
\end{tabular}
  \caption{Label Mapping for dataset ``Republic''}
\label{nlp_joint}
\end{table} 

We see that, the accuracies of  LMs of all the site numbers increase with the increasing 
of training epoch basically. An overfitting occurs at epoch 10 when we use  LM of 6 sites 
as input encoding and 
 LM of 2 sites as output encoding, but it disappear with the increasing of number of sites of output encoding.
The accuracies of  LMs increase with the increasing of sites number.
Even when the number of parameters used in LM is
much less than the one-hot with softmax or one-hot with negative sampling,
the performance of LM is better than one-hot. 

There is an usually used method for big vocabulary
in language model, i.e. the cut-off method: the most 
frequent  words are encoded on-hot, and all other words are common encoded as '$000\cdots 0001$'.
If we view the LM of 6 sites with $\{N_i\}=\{107, 109, 113, 127, 131, 137\}$ 
as a binary encoding, its length is 107+109+113+127+131+137=724. 
We see that, the performance of cut-off method of 724 bits is much lower than the mixed LM of 6 
sites with $\{N_i\}=\{107, 109, 113, 127, 131, 137\}$.

\section{Conclusion}
%

We give an ensemble method so called Label Mapping (LM), 
which  translates a classification problem of huge class number to several classification 
sub-problems of middle class number, and trains a base learner for every sub-problem.
The necessary number of base learners is sub-linear grow with the growing of 
class number.

We propose two design principles, namely, \textbf{Classes high separable} and
\textbf{Base learners independence} of Label Mapping, and give two
classes of Label Mapping and prove they are satisfying the two principles.

As numeric experiments, we show the accuracies
of LM on three datasets, namely, the dataset Cifar-100, the dataset CJK characters and the dataset ``Republic''. 
On all the datasets,
the accuracies of LM increase with the increasing of length. 
When the class number is big (the dataset CJK characters and the dataset ``Republic''), 
specially, the class number
is much greater than the dimension of the last but one layer
of the network, the accuracy of LM plus Network is 
better than the one-hot encoding plus Network with almost same or big number of parameters. 
We compare LM with the classical method ECOC also, the accuracy of LM is much
greater than the accuracy of ECOC of bigger number of parameters.

\vspace{12pt}
\color{red}

\end{document}